\documentclass[fp]{jpsj3}
\usepackage{txfonts}
\usepackage[utf8]{inputenc}
\usepackage{physics}
\usepackage{bm}
\usepackage[dvipdfmx]{graphicx}
\usepackage{subfigure}
\usepackage{color}
\title{Statistical Mechanical Analysis of Catastrophic Forgetting in Continual Learning with Teacher and Student Networks}

\author{Haruka ASANUMA$^1$, Shiro TAKAGI$^1$, Yoshihiro NAGANO$^1$, Yuki YOSHIDA$^1$, Yasuhiko IGARASHI$^{2,3}$, and Masato OKADA$^1$}
\inst{$^1$ 5--1--5 Kashiwanoha, Kashiwa, Chiba, 277--8 \\
$^2$ 1--1--1 Tennoudai, Tsukuba science city, Ibaraki, 30 \\
$^3$ Japan Science and Technology Agency, PRESTO, 4-1-8 Honcho, Kawaguchi, Saitama, 332-0012, Japan} %\\

\abst{
When a computational system continuously learns from an ever-changing environment, it rapidly forgets its past experiences.
This phenomenon is called \textit{catastrophic forgetting}.
While a line of studies has been proposed with respect to avoiding catastrophic forgetting, most of the methods are based on intuitive insights into the phenomenon, and their performances have been evaluated by numerical experiments using benchmark datasets.
Therefore, in this study, we provide the theoretical framework for analyzing catastrophic forgetting by using teacher-student learning. 
Teacher-student learning is a framework in which we introduce two neural networks: one neural network is a target function in supervised learning, and the other is a learning neural network. 
To analyze continual learning in the teacher-student framework, we introduce the similarity of the input distribution and the input-output relationship of the target functions as the similarity of tasks.
In this theoretical framework, we also provide a qualitative understanding of how a single-layer linear learning neural network forgets tasks. 
Based on the analysis, we find that the network can avoid catastrophic forgetting when the similarity among input distributions is small and that of the input-output relationship of the target functions is large.
The analysis also suggests that a system often exhibits a characteristic phenomenon called \textit{overshoot}, which means that even if the learning network has once undergone catastrophic forgetting, it is possible that the network may perform reasonably well after further learning of the current task.
}

\begin{document}
%\linenumbers
\maketitle

\section{Introduction}\label{sec:introduction}
Intelligent systems must continue to adapt to the changing environment to behave suitably in the real world.
An agent needs to remember previously learned experiences while adapting to new knowledge.
Continual learning is a framework to train an agent under a sequence of different tasks to achieve such intelligent behavior \cite{Parisi19, Delange20}.
One of the main challenges in continual learning for neural networks is called \textit{catastrophic forgetting}. A line of studies has reported that when information relevant to the current task is input into a neural network, the knowledge of the previously learned task is suddenly lost \cite{McCloskey89, Ratcliff90, Barnes1959FateOF}.

A variety of methods have been developed to avoid catastrophic forgetting in training a neural network.
The proposed methods can be broadly classified \cite{Parisi19, Delange20} into approaches such as regularization approaches \cite{LwF, Kirkpatrick3521}, dynamic architectural updates \cite{Rusu, Zhou2012}, and memory reply \cite{Rebuffi17, Isele18}.
Most of the previous methods for avoiding catastrophic forgetting are based on intuitive insights into the phenomenon, and their performances have been evaluated by numerical experiments using benchmark datasets.
Therefore, there is room for consideration in the theoretical evaluation of these proposed methods.
If we have a theoretical framework for understanding the phenomenon of catastrophic forgetting, then we can evaluate existing methods with the same framework. 
Additionally, by understanding catastrophic forgetting, we may be able to propose theoretically suitable methods within the framework.
For these reasons, the purpose of this study is to provide a theoretical framework for the phenomenon of catastrophic forgetting.

In this study, we provide the theoretical framework to analyze catastrophic forgetting by using teacher-student learning \cite{SaadSolla95}.
Teacher-student learning is a framework in which we introduce a neural network called \textit{the teacher network} as a target function in supervised learning, and the learning neural network is called \textit{the student network}. The student network learns from the difference between the teacher's output and the student's output.
Teacher-student learning is a method that allows us to analyze the learning dynamics of neural networks from the perspective of statistical mechanics \cite{SaadSolla95, Yoshida19nips, Goldt19nips, Biehl2020}.
For simplicity, we consider analyzing a single-layer linear neural network as the model, which learns two tasks.
We analyze one of the most fundamental learning rules, called stochastic gradient descent (SGD). 

To analyze continual learning in the teacher-student framework, we need to model the similarity between tasks. In this study, we introduce the following two similarity measures: the similarity between input distributions (\textit{input space similarity}) and the similarity between the input-output relationships of teacher networks (\textit{weight space similarity}).
It is believed that the similarity between the input distributions and the input-output relationships of target functions contributes to catastrophic forgetting during continual learning \cite{Goodfellow13, Biehl2020, Bennani20}.
Additionally, since the task in supervised learning is formulated as learning the input-output relationship under a specific input distribution, it is reasonable to introduce these two similarities.
Specifically, we assume that the input distribution is distributed only in a certain subspace of the input space and use the size of the overlap in that subspace as the input space similarity.
We define the weight space similarity as the inner product of the teacher's weight representing the true input-output relationship of each task.

Several researchers \cite{Biehl2020,Bennani20} have theoretically analyzed continual learning.
Biehl et al. \cite{Biehl2020} provided a teacher-student learning framework by using statistical mechanics for analyzing learning dynamics when the task changed continuously.
Our study is different from theirs in that they studied how well student networks adapt to changing tasks, while we focus on the catastrophic forgetting of previous experiences.
Bennani et al. \cite{Bennani20} provided a framework for studying continual learning in the neural tangent kernel (NTK) \cite{Jacot18} regime.
While the framework of Bennani et al. provided upper bounds for generalization error for a wide range of models and inputs, our framework provides an analytical solution for the generalization error for a specific input and model.

We theoretically analyze the abovementioned setup for continual learning.
Based on the analysis, we found that the network can avoid catastrophic forgetting when input space similarity is small and when weight space similarity is large.
In other words, the student network can remember the first task relatively well when the similarity of input distributions is small and when the similarity of the input-output relationship is large.
In addition, a characteristic phenomenon called \textit{overshoot} was observed as a behavior of catastrophic forgetting. Overshoot is the phenomenon in which once an intelligent system learns the current task, it largely forgets the previous task, while continuing to learn the current task makes the student network remember some of the previous forgotten tasks. 
This phenomenon suggests that even if a system undergoes catastrophic forgetting, it may still exhibit better performance if it learns the current task for a longer time.

\section{Model}
In this section, we formulate the continual learning problem with a teacher-student framework \cite{SaadSolla95}, in which a student network aims to learn an input-output relationship realized by a teacher network.
In particular, we focus on the case where the student network learns two tasks in sequential fashion.
We then analyze the qualitative performance of the student network by introducing two types of similarity measures between tasks: \textit{input space similarity} and \textit{weight space similarity}.

\subsection{Teacher-student framework}\label{subsec:two-task-model}
First, we introduce a teacher-student framework.
We consider a regression problem, in which a learning task for the student network is utilized to mimic the true function (teacher network) under a certain input distribution.
In this study, therefore, we introduce two teacher networks as target functions and two input distributions.
\begin{figure}
\centering
\includegraphics[width = \linewidth]{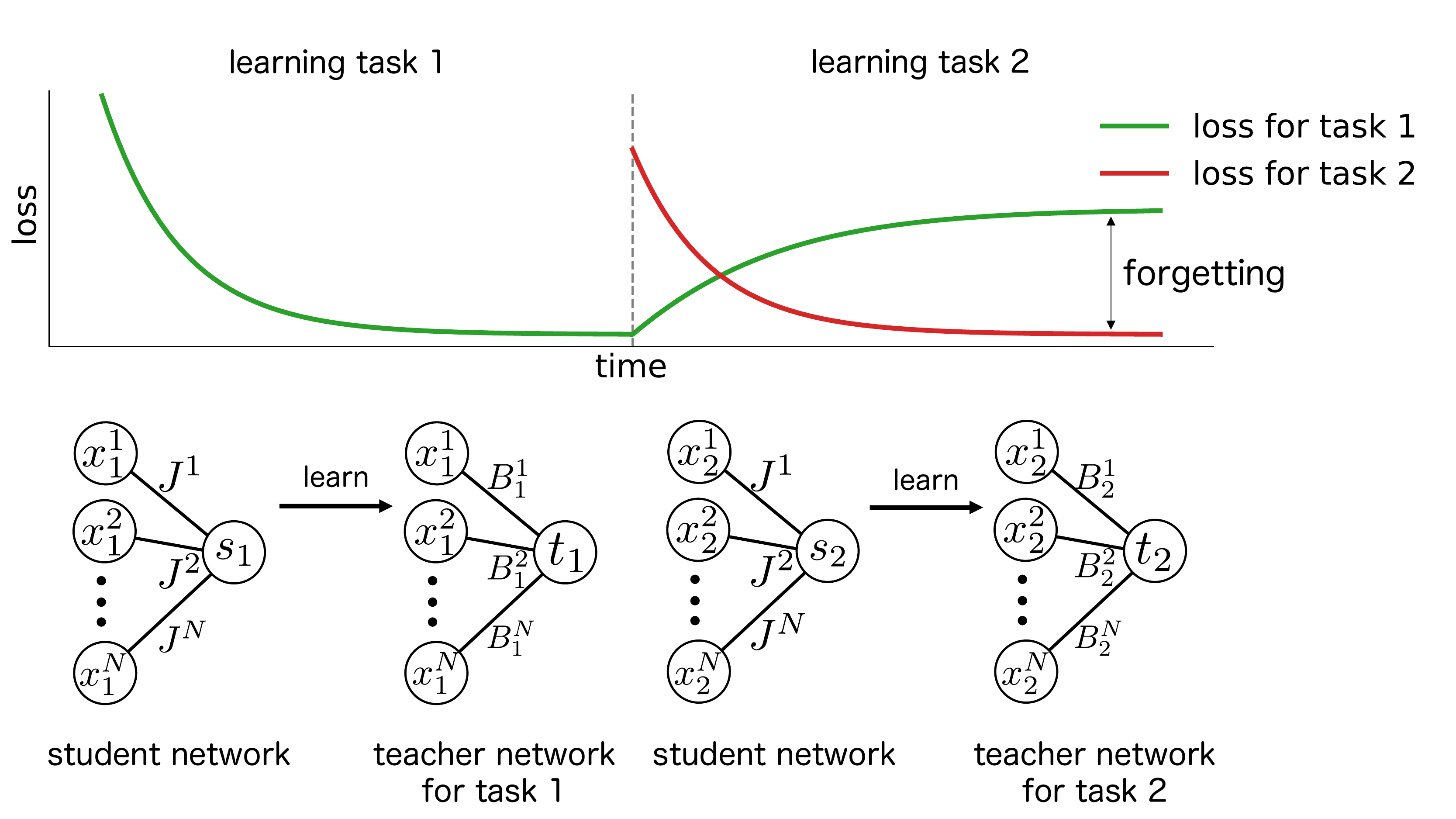}
\caption{Schematic diagram of the continual learning problem in a teacher-student setup. The student network learns task 1 first and then learns task 2. The upper figure shows the ideal behavior of the generalization error for tasks 1 and 2.}
\label{fig:2task_learning}
\end{figure}

We consider that a single-layer linear neural network learns two tasks sequentially, as illustrated in Fig. \ref{fig:2task_learning}.
The student network receives $N$-dimensional input data $\bm{x}_v \in \mathbb{R}^N, N \in \mathbb{N}$ and calculates output $s_v = \bm{J}^T\bm{x}_v \in \mathbb{R}$, where $\bm{J} \in \mathbb{R}^N$ is a weight that is optimized while learning.
The subscript of each variable (e.g., $\bm{x}_v, v \in \left\{1, 2\right\}$) indicates the task index.
The learning task for the student network is to mimic the output of a teacher network $t_v = \bm{B}_v^T \bm{x}_v$ under a certain input distribution $\bm{x}_v \sim P(\bm{x}_v)$.
We refer to task 1 as a pair of teacher weight and input distribution $(\bm{B}_1, P(\bm{x}_1))$ and task 2 as a pair $(\bm{B}_2, P(\bm{x}_2))$.
For simplicity, we assume that each element of input data $x_v^i$ is independently sampled from the normal distribution, $P(\bm{x}_v) = \mathcal{N}(\bm{x}_v |0, \sigma_v^2I_v)$, where $I_v$ characterizes the data distributions of $P(\bm{x}_v)$.
We will define $I_v$ in detail in Section \ref{subsec:input-space-similarity}.

In this study, we focus on the on-line learning setting.
In the on-line framework, every time input data $\bm{x}_v$ are given, we update student weight $\bm{J}$: input data $\bm{x}_v$ will never again be used for learning.
Since the input used for learning is discarded and the previous and next inputs are statistically independent, student weight $\bm{J}$
is independent of new input data $\bm{x}_v$.
We update student weight $\bm{J}$ to minimize the error between its output and the output of the teacher network.
We use the squared error function $\epsilon_v = \frac{1}{2}(t_v - s_v)^2$, which is most commonly used for regression.
To optimize the student's weights, we use stochastic gradient descent (SGD).
The update amount of the weights with SGD while learning task 1 is written as follows:
\begin{align}
\Delta \bm{J} &= -\frac{\eta}{N}\dv[]{\epsilon_1}{\bm{J}} = \frac{\eta}{N}\bm{x}_1(t_1 - s_1)\\
&= \frac{\eta}{N}\bm{x}_1^T(\bm{B}_1^T\bm{x}_1 - \bm{J}^T\bm{x}_1),\label{eq:update_for_task1}
\end{align}
and the update amount of the weights with SGD while learning task 2 is written as follows:
\begin{align}
\Delta \bm{J} &= -\frac{\eta}{N}\dv[]{\epsilon_2}{\bm{J}} = \frac{\eta}{N}\bm{x}_2(t_2 - s_2)\\
&= \frac{\eta}{N}\bm{x}_1^T(\bm{B}_2^T\bm{x}_2 - \bm{J}^T\bm{x}_2),\label{eq:update_for_task2}
\end{align}
in which we set the learning rate as $\frac{\eta}{N}$ so that our macroscopic system is $N$-independent.
For simplicity of calculation, we assume that the weights of the initial student network $\bm{J}_0$ and teacher networks $\bm{B}_1$ and $\bm{B}_2$ are sampled independently from the normal distribution: $\mathcal{N}(J_0^i|0, \frac{\sigma_{J}^2}{N})$, $\mathcal{N}(B_1^i|0, \frac{\sigma_{B1}^2}{N})$, and $\mathcal{N}(B_2^i|0, \frac{\sigma_{B2}^2}{N})$.

To evaluate model performance, we focus on a metric called generalization error for each task.
The generalization errors for task 1 ($\epsilon_{g1}$) and task 2 ($\epsilon_{g2}$) are defined as follows:
\begin{align}
\epsilon_{g1}
&=\int d\bm{x}_1\epsilon_1P(\bm{x}_1)=\int d\bm{x}_1\frac{1}{2}(t_1 - s_1)^2P(\bm{x}_1),
\label{eq:learning_dynamics_of_task1}\\
\epsilon_{g2}
&=\int d\bm{x}_2\epsilon_2P(\bm{x}_2)=\int d\bm{x}_2\frac{1}{2}(t_2 - s_2)^2P(\bm{x}_2).
\label{eq:learning_dynamics_of_task2}
\end{align}

\subsection{Input space similarity}\label{subsec:input-space-similarity}
\begin{figure}
\centering
\includegraphics[width = 0.5\linewidth]{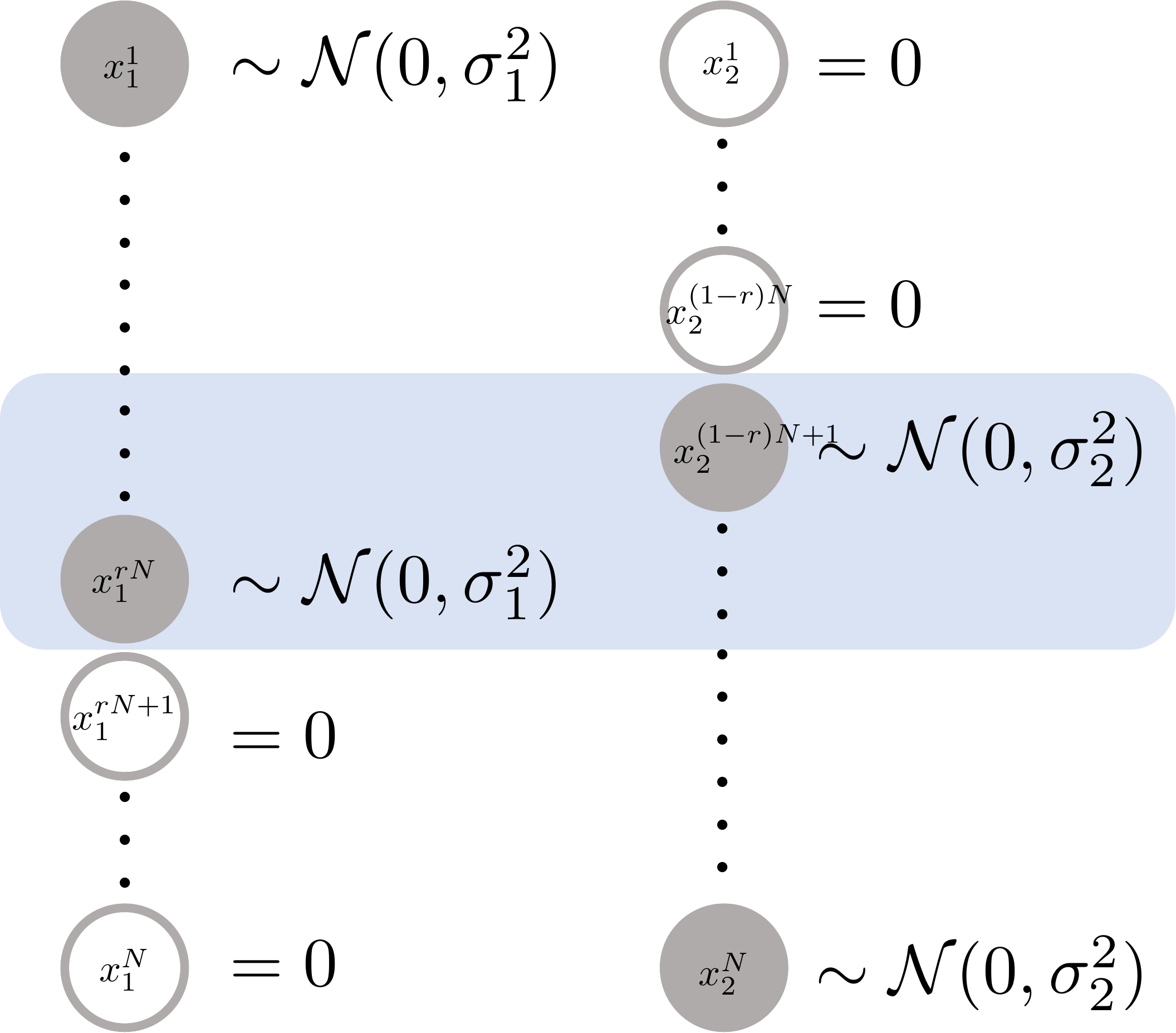}
\caption{Schematic diagram of input distributions for two tasks and input space similarity $r$. Gray nodes have nonzero values and white nodes correspond to zero. The range enclosed in blue has nonzero values in common with $\bm{x}_1$ and $\bm{x}_2$.
Since $\{x_1^i, x_2^i\mid i = (1 - r)N + 1, \dots, rN\}$ have values in common, we can see that there are $(2r - 1) N$ dimensions for common nonzero input space. Therefore, by determining $r$, we can determine the ratio of the common space, $(2r - 1)$.}
\label{fig:input_explain}
\end{figure}

In this study, as discussed in Section \ref{subsec:two-task-model},
we introduced two input distributions, $P(\bm{x}_1)$ and $P(\bm{x}_2)$. In this section, we model these input distributions to evaluate the relationship between inputs for two tasks.
We characterize the difference between $P(\bm{x}_1)$ and $P(\bm{x}_2)$ by the difference in the input space, that is, $I_1$ and $I_2$ \textcolor{black}{which were defined in \ref{subsec:two-task-model}.}
Here, we introduce a parameter, $r$, as the proportion of the subspace in which each input has a nonzero value from the total input space \textcolor{black}{to define $I_1$ and $I_2$ in a simple way}. \textcolor{black}{As we'll see in more detail later, setting $r$ determines the proportion of the common space of $\bm{x}_1$ and $\bm{x}_2$}. 
For simplicity of calculation, we set $I_1 = I_{1:rN}$ and $I_2 = I_{(1 - r)N:N}$. 
$I_{i:j}$ indicates a matrix in which the values of diagonal components from the $i$-th to the $j$-th dimensions are ones, and the rest are zeros.
\textcolor{black}{Fig. \ref{fig:input_explain} is a schematic diagram of the input space of task 1 and task 2. Gray nodes have nonzero values and white nodes correspond to zero. The range enclosed in blue has nonzero values in common with $\bm{x}_1$ and $\bm{x}_2$.
Since $\{x_1^i, x_2^i\mid i = (1 - r)N + 1, \dots, rN\}$ have values in common, we can see that there are $(2r - 1) N$ dimensions for common nonzero input space. Therefore, by determining $r$, we can determine the ratio of the common space, $(2r - 1)$.}
In other words, we assume that the first $rN$ input elements for task 1 $\{x_1^i\mid i = 1,\ldots, rN\}$ are independently sampled from the normal distribution $\mathcal{N}(x_1^i\mid 0, \sigma_1^2)$, as discussed in Section \ref{subsec:two-task-model}, and the last $(1 - r)N$ input elements for task 1 $\{x_1^i\mid i = rN + 1 ,\ldots, N\}$ equal zero. We also assume that the first $(1 - r)N$ input elements for task 2 $\{x_2^i\mid i = 1,\ldots, (1 - r) N\}$ equal zero and that the last $rN$ input elements for task 2 $\{x_2^i\mid i = (1 - r) N + 1,\ldots, N\}$ are independently sampled from the normal distribution $\mathcal{N}(x_2^i\mid 0, \sigma_2^2)$, as discussed in Section \ref{subsec:two-task-model}.
We can see that there are $(2r - 1) N$ dimensions for common nonzero input space. Therefore, as mentioned above, determining $r$ is equivalent to the ratio of common space.
We assume the common input space and $0.5\le r \le 1$. 

In this study, we modeled the differences in the data distribution by the difference in the input subspace. Here, we explain how this reflects a practical situation by using an MNIST \cite{MNIST} dataset as a motivating example.
The MNIST dataset contains white pixels with zero input and colored pixels with nonzero input.
The colored pixels in the dataset of the number ``3'' and those of the number ``8'' are assumed to have a large number of colored pixels in common because of the similarity in their shapes. In other words, the ratio $r$ of the common space is close to 1.
Moreover, the colored pixels in the dataset of the number ``1'' and those of the number ``8'' are not similar, so the number of colored pixels in common is estimated to be small. In such a situation, the ratio $r$ of the common space is close to 0.5.
Based on these observations, the proposed modeling of the data distribution is reasonable and reflects a real-world situation.

\subsection{Weight space similarity}
As shown in Section \ref{subsec:two-task-model}, a task is characterized by the input distribution and a teacher network. Therefore, the similarity between tasks will be determined by the input space similarity and the similarity of the teachers.
Since input space similarity was defined in Section \ref{subsec:input-space-similarity}, we will define the similarity of teachers in this section.

In this study, model learning task 1 is equivalent to student weight $\bm{J}$ approaching teacher weight $\bm{B}_1$ under the data distribution $P(\bm{x}_1)$, and model learning task 2 is equivalent to student weight $\bm{J}$ approaching teacher weight $\bm{B}_2$ under the data distribution $P(\bm{x}_2)$.
Therefore, we define the similarity of teachers by using $\bm{B}_1$ and $\bm{B}_2$ as follows:
\begin{equation}
q \coloneqq \frac{\bm{B}_1^T\bm{B}_2}{\|\bm{B}_1\| \|\bm{B}_2\|}.
\end{equation}
we call the parameter $q$ \textit{weight space similarity}.
When there is no correlation at all between tasks 1 and 2, $q = 0$, and when tasks 1 and 2 are completely similar, $q = 1$.

\section{Theory}
In this section, we show the analytical solution for the generalization error under a continual learning setting.
The generalization errors shown in \eqref{eq:learning_dynamics_of_task1} and \eqref{eq:learning_dynamics_of_task2} can be rewritten as follows:
\begin{align}
\epsilon_{g1}
&= \frac{1}{2}\langle\bm{B}_1^T\bm{x}_1\bm{x}_1^T\bm{B}_1\rangle -  \langle\bm{B}_1^T\bm{x}_1\bm{x}_1^T\bm{J}\rangle + \frac{1}{2}\langle\bm{J}^T\bm{x}_1\bm{x}_1^T\bm{J}\rangle
\label{eq:learning_dynamics_of_task1_expand}\\
\epsilon_{g2}
&=\frac{1}{2}\langle\bm{B}_2^T\bm{x}_2\bm{x}_2^T\bm{B}_2\rangle -  \langle\bm{B}_2^T\bm{x}_2\bm{x}_2^T\bm{J}\rangle + \frac{1}{2}\langle\bm{J}^T\bm{x}_2\bm{x}_2^T\bm{J}\rangle
\label{eq:learning_dynamics_of_task2_expand}
\end{align}
The angle bracket $\langle \cdot \rangle$ expresses the expectation for input distribution.
Here, we introduce \textit{order parameters} to capture the state of the system macroscopically:
$\sigma_v^2Q_v = \langle\bm{J}^T \bm{x}_v \bm{x}_v^T\bm{J}\rangle = \bm{J}^T\langle \bm{x}_v \bm{x}_v^T\rangle\bm{J}$, $\sigma_v^2R_v^u = \langle\bm{B}_u^T \bm{x}_v \bm{x}_v^T\bm{J}\rangle = \bm{B}_u^T\langle \bm{x}_v \bm{x}_v^T\rangle\bm{J}$ and $\sigma_v^2T_v^u \coloneqq \langle\bm{B}_u\bm{x}_v \bm{x}_v^T\bm{B}_u\rangle = \bm{B}_u\langle\bm{x}_v \bm{x}_v^T\rangle\bm{B}_u$ using task number $u, v \in \{1, 2\}$, that is 
\begin{align}
\sigma_1^2 Q_1 \coloneqq \bm{J}^T\langle \bm{x}_1\bm{x}_1^T\rangle\bm{J},\\
\sigma_2^2 Q_2 \coloneqq \bm{J}^T\langle \bm{x}_2\bm{x}_2^T\rangle\bm{J},\\ 
\sigma_1^2\sigma_2^2 Q_{12} \coloneqq \bm{J}^T\langle \bm{x}_1\bm{x}_1^T\rangle\langle \bm{x}_2\bm{x}_2^T\rangle\bm{J},\\
\sigma_1^2 R_1^1 \coloneqq \bm{B}_1^T\langle \bm{x}_1\bm{x}_1^T\rangle\bm{J},\\
\sigma_2^2 R_2^1 \coloneqq \bm{B}_1^T\langle \bm{x}_2\bm{x}_2^T\rangle\bm{J},\\
\sigma_1^2 R_2^1 \coloneqq \bm{B}_2^T\langle \bm{x}_1\bm{x}_1^T\rangle\bm{J},\\
\sigma_2^2 R_2^2 \coloneqq \bm{B}_2^T\langle \bm{x}_2\bm{x}_2^T\rangle\bm{J},\\ \sigma_1^2\sigma_2^2 R_{12}^1 \coloneqq \bm{B}_1^T\langle \bm{x}_1\bm{x}_1^T\rangle\langle \bm{x}_2\bm{x}_2^T\rangle\bm{J},\\
\sigma_1^2\sigma_2^2 R_{12}^2 \coloneqq \bm{B}_2^T\langle \bm{x}_1\bm{x}_1^T\rangle\langle \bm{x}_2\bm{x}_2^T\rangle\bm{J}, \\
\sigma_1^2T_1^1 \coloneqq \bm{B}_1^T\langle \bm{x}_1\bm{x}_1^T\rangle\bm{B}_1,\\ 
\sigma_2^2T_2^2 \coloneqq \bm{B}_2^T\langle \bm{x}_2\bm{x}_2^T\rangle\bm{B}_2,\\ \sigma_1^2\sigma_2^2T_{12}^1 \coloneqq \bm{B}_1^T\langle \bm{x}_1\bm{x}_1^T\rangle\langle \bm{x}_2\bm{x}_2^T\rangle\bm{B}_1,\\
\sigma_1^2\sigma_2^2T_{12}^2 \coloneqq \bm{B}_2^T\langle \bm{x}_1\bm{x}_1^T\rangle\langle \bm{x}_2\bm{x}_2^T\rangle\bm{B}_2,
\end{align}
so
\begin{align}
Q_1 \coloneqq \bm{J}^TI_1\bm{J},\label{eq:define_q1}\\
Q_2 \coloneqq \bm{J}^TI_2\bm{J},\label{eq:define_q2}\\ 
Q_{12} \coloneqq \bm{J}^TI_1I_2\bm{J},\label{eq:define_q12}\\
R_1^1 \coloneqq \bm{B}_1^TI_1\bm{J},\label{eq:define_r11}\\
R_2^1 \coloneqq \bm{B}_1^TI_2\bm{J},\label{eq:define_r21}\\
R_1^2 \coloneqq \bm{B}_2^TI_1\bm{J},\label{eq:define_r12}\\
R_2^2 \coloneqq \bm{B}_2^TI_2\bm{J},\label{eq:define_r22}\\ 
R_{12}^1 \coloneqq \bm{B}_1^TI_1I_2\bm{J},\label{eq:define_r121}\\
R_{12}^2 \coloneqq \bm{B}_2^TI_1I_2\bm{J},\label{eq:define_r122} \\
T_1^1 \coloneqq \bm{B}_1^TI_1\bm{B}_1,\label{eq:define_t11}\\ 
T_2^2 \coloneqq \bm{B}_2^TI_2\bm{B}_2,\label{eq:define_t22}\\ 
T_{12}^1 \coloneqq \bm{B}_1^TI_1I_2\bm{B}_1,\label{eq:define_t121}\\
T_{12}^2 \coloneqq \bm{B}_2^TI_1I_2\bm{B}_2\label{eq:define_t122}.
\end{align}
By using these definitions, we can rewrite \eqref{eq:learning_dynamics_of_task1_expand} and \eqref{eq:learning_dynamics_of_task2_expand} as follows:
\begin{align}
\epsilon_{g1}&= \frac{\sigma_1^2}{2}(Q_1 - 2 R_1^1 + T_1^1),\\
\epsilon_{g2}&= \frac{\sigma_2^2}{2}(Q_2 - 2 R_2^2 + T_2^2).
\end{align}
The values of $Q_v$ and $R_v^u$ change while learning, and the value of $T_v^u$ is constant because $\bm{J}$ changes and $\bm{B}_v$ does not change while learning.
Therefore, when the dynamics of $Q_v$ and $R_v^u$ are calculated, we can calculate the dynamics of the generalization error and quantitatively evaluate model performance.
In a standard supervised learning setting, it is known from previous studies that the dynamics of order parameters can be derived analytically by taking the limit of $N\rightarrow\infty$ for the number of input dimensions $N$\cite{SaadSolla95, Yoshida19nips, Goldt19nips}.
This operation is inspired by the thermodynamic limit, which is the limit of the number of particles in thermodynamics.

As mentioned in Section \ref{sec:introduction}, it is known that learning task 2 after learning task 1 causes \textit{catastrophic forgetting}, in which the student network rapidly forgets task 1.
We note that the generalization error $\epsilon_{g1}$ of task 1 \textbf{while learning task 2} expresses the amount of forgetting.  
Therefore, we analytically derive the following three generalization errors: 
\begin{itemize}
    \item the generalization error of task 1 while learning task 1, 
    \item the generalization error of task 2 while learning task 2, 
    \item and the generalization error of task 1 while learning task 2. 
\end{itemize}

\subsection{Task 1 learning}\label{subsec:task-1-learning}
Let the update amount of $Q_1$ be $\Delta Q_1$ and the update amount of $R_1$ be $\Delta R_1$. Here, $\Delta Q_1$ means that $(\bm{J} + \Delta\bm{J})^TI_1(\bm{J} + \Delta\bm{J}) - \bm{J}^TI_1\bm{J}$, and $\Delta R_1^1$ means that $\bm{B}_1^TI_1(\bm{J} + \Delta\bm{J} - \bm{B}_1^TI_1\bm{J})$.
From the update amount of $\bm{J}$ defined in \eqref{eq:update_for_task1}, $\Delta Q_1$ and $\Delta R_1^1$ while learning task 1 can be written as follows:
\begin{align}
\Delta Q_1 &= (\bm{J} + \Delta\bm{J})^TI_1(\bm{J} + \Delta\bm{J}) - \bm{J}^TI_1\bm{J}\\
&= \Delta\bm{J}^TI_1\bm{J} + \bm{J}I_1\Delta\bm{J} + \Delta\bm{J}^TI_1\Delta\bm{J}\\
&= 2\frac{\eta}{N}\{\bm{x}_1(\bm{B}_1^T\bm{x}_1 - \bm{J}^T\bm{x}_1)\}^TI_1\bm{J} + \frac{\eta^2}{N^2}\{\bm{x}_1(\bm{B}_1^T\bm{x}_1 - \bm{J}^T\bm{x}_1)\}^TI_1\bm{x}_1(\bm{B}_1^T\bm{x}_1 - \bm{J}^T\bm{x}_1)\label{eq:update_Q_1_task1}\\
\Delta R_1^1 &= \bm{B}_1^T I_1(\bm{J} + \Delta\bm{J}) - \bm{B}_1^T I_1\bm{J} = \bm{B}_1I_1\Delta\bm{J}\\
&= \frac{\eta}{N}\bm{B}_1^T I_1 \bm{x}_1(\bm{B}_1^T\bm{x}_1 - \bm{J}^T\bm{x}_1).\label{eq:update_R_1^1_task1}
\end{align}
By taking the limit of $N\rightarrow\infty$ for the number of input dimensions (thermodynamic limit), we can replace the difference equations with differential equations:
\begin{align}
\dv[]{ Q_1}{m} &=2\frac{\eta}{N}\{\bm{B}_1^T\langle\bm{x}_1\bm{x}_1^T\rangle\bm{J} - \bm{J}^T\langle\bm{x}_1\bm{x}_1^T\rangle\bm{J}\} \\
&+ \frac{\eta^2}{N^2}\{\bm{B}_1^T\langle\bm{x}_1\bm{x}_1^T\bm{x}_1\bm{x}_1^T\rangle\bm{B}_1 - 2 \bm{B}_1^T\langle\bm{x}_1\bm{x}_1^T\bm{x}_1\bm{x}_1^T\rangle\bm{J}^T + \bm{J}_1^T\langle\bm{x}_1\bm{x}_1^T\bm{x}_1\bm{x}_1^T\rangle\bm{J}\}\\
&= 2\frac{\eta\sigma_1^2}{N} (R_1^1 - Q_1) + \frac{\eta^2r\sigma_1^4}{N}(T_1^1 + Q_1 - 2R^1_1),\\
\dv[]{R_1^1}{m}  &= \frac{\eta}{N}\{\bm{B}_1^T \langle \bm{x}_1^T\bm{x}_1\rangle\bm{B}_1 - \bm{B}_1^T \langle \bm{x}_1^T\bm{x}_1\rangle\bm{J}\}\\
&=\frac{\eta\sigma_1^2}{N}(T_1^1 - R_1^1),\\
\dv{Q_1}{t} &= \eta^2 r^2\sigma_1^4(T_1^1 + Q_1 - 2 R_1^1) + 2\eta r\sigma_1^2(R_1^1 - Q_1),\\
\dv{R_1^1}{t} &= \eta\sigma_1^2 r (T_1^1 - R_1^1).
\end{align}
In these equations, $t\coloneqq \frac{m}{rN}$ 
represents time (normalized number of steps).
We can solve the differential equations as follows: 
\begin{align}
R_1^1(t) &= r\sigma^2_{B1} \Bigl(1 - \exp(-\eta r\sigma_1^2t)\Bigr),\label{eq:r1^1_task1}\\
Q_1(t) &= r(\sigma_J^2 + \sigma_{B1}^2)\exp\left(-t(\eta r\sigma_1^2(2 - \eta r\sigma_1^2))\right)- 2r\sigma_{B1}^2\exp\left(-\eta r\sigma_1^2t\right) + r\sigma_{B1}^2.\label{eq:q1_task1}
\end{align}
We used the initial value of $R_1^1(t)$ as $R_1^1(0) = \bm{B}_1^TI_1\bm{J} = \sum_{i  =1}^{rN}B_1^iJ^i = 0$, the initial value of $T_1^1$ as $T_1^1 = \bm{B}_1I_1\bm{B}_1 = \sum_{i = 1}^{rN}B_1^iB_1^i = r\sigma_{B1}^2$ and the initial value of $Q_1^1(t)$ as $Q_1^1(0) = \bm{J}^TI_1\bm{J} = \sum_{i = 1}^{rN}J^iJ^i = r\sigma_J^2$.
By using the solutions of $Q_1$ and $R_1^1$, we obtain the dynamics of the generalization error $\epsilon_{g1}$ of task 1 while learning task 1 as follows:
\begin{align}
\epsilon_{g1}
&=\frac{\sigma_1^2}{2}(T_1^1 - 2R_1^1 + Q_1 )\\
&= \frac{\sigma_1^2}{2}r(\sigma_{B1}^2 + \sigma_J^2)\exp(-t\eta r \sigma_1^2(2 - \eta r \sigma_1^2)).
\label{eq:eg1_task1}
\end{align}
$R_1^1(t)$ and $Q_1(t)$ are substituted by \eqref{eq:r1^1_task1} and \eqref{eq:q1_task1}.

\subsection{Task 2 learning}
Once the student network has learned task 1 sufficiently and $\epsilon_{g1}$ is sufficiently close to $0$, the student network learns task 2. 
\textcolor{black}{
This means that the output of the student network matches the output of the teacher network, so we assume that $B_1^i = J^i$ when the indexes of $\bm{B}_1$ and $\bm{J}$ are in the range $1\le i \le rN$.
By making this assumption, the analytical solution of $\epsilon_{g1}$ and $\epsilon_{g2}$ can be obtained accurately.
}
First, we calculate the generalization error, $\epsilon_{g2}$, of task 2 while learning task 2.
Equation \eqref{eq:update_for_task2} provides the update rule of order parameters $Q_2$ and $R_2^2$ while learning task 2 in the form of difference equations:
\begin{align}
\Delta Q_2 &= (\bm{J} + \Delta\bm{J})^TI_2(\bm{J} + \Delta\bm{J}) - \bm{J}^TI_2\bm{J}\\
&= \Delta\bm{J}^TI_2\bm{J} + \bm{J}I_2\Delta\bm{J} + \Delta\bm{J}^TI_2\Delta\bm{J}\\
&=\frac{2\eta}{N}\{\bm{x}_2(\bm{B}_2^T\bm{x}_2 - \bm{J}^T\bm{x}_2)\}^TI_2\bm{J} + \frac{\eta^2}{N^2}\{\bm{x}_2(\bm{B}_2^T\bm{x}_2 - \bm{J}^T\bm{x}_2)\}^TI_2\bm{x}_2(\bm{B}_2^T\bm{x}_2 - \bm{J}^T\bm{x}_2),\label{eq:update_Q_2_task2}\\
\Delta R_2^2 &= \bm{B}_2^T I_2(\bm{J} + \Delta\bm{J}) - \bm{B}_2^T I_2\bm{J} = \bm{B}_2^TI_2\Delta\bm{J}\\
&= \frac{\eta}{N}\bm{B}_2^TI_2\bm{x}_2(\bm{B}_2^T\bm{x}_2 - \bm{J}^T\bm{x}_2).\label{eq:update_Q_2^2_task2}
\end{align}
Likewise, in Section \ref{subsec:task-1-learning}, we can replace the difference equations with the differential equations by taking the expectation for input $\bm{x}_2$:
\begin{align}
\dv[]{Q_2}{m} 
&=\frac{\eta^2r\sigma_2^4}{N}(T_2^2 + Q_2 - 2R^2_2) + 2\frac{\eta\sigma_2^2}{N} (R_2^2 - Q_2),\\
\dv[]{R_2^2}{m} &=\frac{\eta\sigma_2^2}{N}(T_2^2 - R_2^2),\\
\dv{Q_2}{t} &= \eta^2 r^2\sigma_2^4(T_2^2 + Q_2 - 2 R_2^2) + 2\eta r\sigma_2^2(R_2^2 - Q_2),\\
\dv{R_2^2}{t} &= \eta\sigma_2^2 r (T_2^2 - R_2^2).
\end{align}
We can solve these differential equations as follows:
\begin{align}
Q_2(t) &=\left(r\sigma_{B2}^2 + (2r - 1)\sigma_{B1}^2 + (1- r\sigma_J^2)\right.
 - 2q(2r - 1)N\sigma_{B1}\sigma_{B2}\Bigr)\exp(-t\eta r\sigma_2^2 (2 -\eta r\sigma_2^2))\nonumber\\
 &+ 2(q(2r - 1)\sigma_{B1}\sigma_{B2} - r\sigma_{B2}^2)\exp(-\eta\sigma_2^2t) + r\sigma_{B2}^2,\label{eq:q2_task2}\\
R_2^2(t) &= r\sigma_{B2}^2 + \left(q(2r - 1)\sigma_{B1}\sigma_{B2} - r\sigma_{B2}^2\right)\exp\left(-\eta r\sigma_2^2t\right).\label{eq:r2^2_task2}
\end{align}
Assuming that task 1 has been sufficiently trained and $\epsilon_{g1}$ is sufficiently asymptotic to ``0,'' \textcolor{black}{we consider $B_1^i = J^i$ when the indexes of $\bm{B}_1$ and $\bm{J}$ are in the range $1 \le i\le rN$.} 
So we set the initial value of $R_2^2(t)$ as $R_2^2(0) = \bm{B}_2^TI_2\bm{J} = \sum_{i  =(1-r)N}^{rN}B_2^iB_1^i + \sum_{i = rN}^N B_2^iJ^i = (2r - 1)q\sigma_{B1}\sigma_{B2}$ and the initial value of $Q_2(t)$ as $Q_2(0) =
\bm{J}^TI_2\bm{J} = \sum_{i = (1 - r)N}^{rN}B_1^iB_1^i + \sum_{i = rN}^NJ^iJ^i = (2r - 1)\sigma_{B1}^2 + (1 - r)\sigma_J^2$.
By using the solutions of $Q_2$ and $R_2^2$, we obtain the dynamics of the generalization error of task 2 while learning task 2 as follows:
\begin{align}
\epsilon_{g2}
&=\frac{\sigma_2^2}{2}(T_2^2 - 2R_2^2 + Q_2 )\\
&=\frac{\sigma_2^2}{2}\left(r\sigma_{B2}^2 + (1 - r)\sigma_J^2 + (2r - 1)\right.\sigma_{B1}^2\nonumber\\
 &- 2(2r - 1)\sigma_{B1}\sigma_{B2}q\Bigr)\exp(-t\eta r\sigma_2^2 (2 - \eta r\sigma_2^2)).\label{eq:eg2_task2}
\end{align}
$R_2^2(t)$ and $Q_2(t)$ are substituted by \eqref{eq:r2^2_task2} and \eqref{eq:q2_task2}.

Next, we derive the generalization error, $\epsilon_{g1}$, of task 1 while learning task 2.
Equation \eqref{eq:update_for_task2} provides the update rule of order parameters $Q_1$ and $R_1^1$ while learning task 2 in the form of difference equations:
\begin{align}
\Delta Q_1 &= (\bm{J} + \Delta\bm{J})^TI_1(\bm{J} + \Delta\bm{J}) - \bm{J}^TI_1\bm{J}\\
&= \Delta\bm{J}^TI_1\bm{J} + \bm{J}I_1\Delta\bm{J} + \Delta\bm{J}^TI_1\Delta\bm{J}\\
&=\frac{2\eta}{N}\{\bm{x}_2(\bm{B}_2^T\bm{x}_2 - \bm{J}^T\bm{x}_2)\}^TI_1\bm{J} + \frac{\eta^2}{N^2}\{\bm{x}_2(\bm{B}_2^T\bm{x}_2 - \bm{J}^T\bm{x}_2)\}^TI_1\bm{x}_2(\bm{B}_2^T\bm{x}_2 - \bm{J}^T\bm{x}_2),\label{eq:update_Q_1_task2}\\
\Delta R_1^1 &= \bm{B}_1^T I_1(\bm{J} + \Delta\bm{J}) - \bm{B}_1^T I_1\bm{J} = \bm{B}_1I_1\Delta\bm{J}\\
&= \frac{\eta}{N}\bm{B}_1I_1\bm{x}_2(\bm{B}_2^T\bm{x}_2 - \bm{J}^T\bm{x}_2).\label{eq:update_R_1^1_task2}
\end{align}
Likewise, in Section \ref{subsec:task-1-learning}, we can replace the difference equations with the differential equations by taking the expectation for inputs $\bm{x}_1$ and $\bm{x}_2$:
\begin{align}
\dv[]{Q_1}{m} 
&=\frac{\eta^2r\sigma_2^4}{N}(T_{12}^2 + Q_{12} - 2R^2_{12}) + 2\frac{\eta\sigma_2^2}{N} (R_{12}^2 - Q_{12}), \label{eq:update_Q1}\\
\dv[]{R_1^1}{m} &=\frac{\eta\sigma_2^2}{N}(q^\prime - R_{12}^1),\label{eq:update_R1^1}
\end{align}
where $q^\prime = \bm{B}_1I_1I_2\bm{B}_2$. To derive the dynamics of $Q_1$ and $R_1^1$ while learning task 2, we must derive the dynamics of $Q_{12}$, $R_{12}^1$, and $R_{12}^2$.
We can derive the difference equation as follows by performing the same calculations as in \eqref{eq:update_Q1} and \eqref{eq:update_R1^1}:
\begin{align}
\dv[]{R_{12}^1}{m} &= \frac{\eta \sigma_2^2}{N}(q^\prime - R_{12}^1),\\
\dv[]{R_{12}^2}{m}
&=\frac{\eta \sigma_2^2}{N}(T_{12}^2 - R_{12}^2),\\
\dv[]{Q_{12}}{m} 
&= \frac{2\eta\sigma_2^2}{N}(R_{12}^2 - Q_{12})
+ \frac{\eta^2\sigma_2^4(2r -1)}{N}(T_2^2 - 2R_{2}^2 + Q_2).
\end{align}
Therefore, the above equations can be written as follows:
\begin{align}
\dv{R_1^1}{t} &= \eta r\sigma_2^2(q^\prime - R_{12}^1),\\
\dv{Q_{1}}{t}&= 2\eta r\sigma_2^2(R_{12}^2 - Q_{12})
+ \eta^2\sigma_2^4r(2r -1)(T_2^2 - 2R_{2}^2 + Q_2),\\
\dv{ R_{12}^1}{t} &= \eta r\sigma_2^2 (q^\prime - R_{12}^1), \\
\dv{ R_{12}^2}{t} 
&=\eta r\sigma_2^2(T_{12}^2 - R_{12}^2),\\
\dv{Q_{12}} {t}&= 2\eta r\sigma_2^2(R_{12}^2 - Q_{12})
+ \eta^2\sigma_2^4r(2r -1)(T_2^2 - 2R_{2}^2 + Q_2).
\end{align}
Solving these differential equations yields the following solutions:
\begin{align}
R_{12}^{1}(t) &= 
q(2r - 1)\sigma_{B1}\sigma_{B2}
+ (2r - 1)(\sigma_{B1}^2 - q\sigma_{B1}\sigma_{B2})\exp\left(-t\eta r\sigma_2^2\right) ,\\
R_{12}^{2} (t) &= 
(2r - 1)\Bigl\{\sigma_{B2}^2
+ (q\sigma_{B1}\sigma_{B2} - \sigma_{B2}^2)\exp(-t\eta r\sigma_2^2)\Bigr\},\\
Q_{12}(t) &=2(2r - 1)(q\sigma_{B1}\sigma_{B2} - \sigma_{B1}^2)\exp(-t\eta r\sigma_2^2)\nonumber\\
&+C_2\exp(-t\eta r\sigma_2^2(2 - \eta r\sigma_2^2))
+\Bigl\{C_1 - C_2\Bigr\}\exp(-2t\eta r\sigma_2^2) + (2r - 1)\sigma_{B2}^2,\\
R_1^1(t) &= (2r - 1)(\sigma_{B1}^2 - q\sigma_{B1}\sigma_{B2})\exp(-t\eta r\sigma_2^2)\\
&+ r\sigma_{B1}^2 - (2r - 1)(\sigma_{B1}^2 - q\sigma_{B1}\sigma_{B2}),\label{eq:r1^1_task2}\\
Q_1(t) &= 2(2r - 1)(q\sigma_{B1}\sigma_{B2} - \sigma_{B1}^2)\exp(-t\eta r\sigma_2^2)\nonumber\\
&+C_2\exp(-t\eta r\sigma_2^2(2 - \eta r\sigma_2^2))+\Bigl\{C_1-C_2\Bigr\}\exp(-2t\eta r\sigma_2^2)\nonumber\\
&+(2r - 1)(\sigma_{B2}^2 - \sigma_{B1}^2) + r\sigma_{B1}^2,\\
\label{eq:q1_task2}
C_1 &= (2r - 1)(\sigma_{B1}^2 + \sigma_{B2}^2 - 2q\sigma_{B1}\sigma_{B2}),\\
C_2 &= \frac{2r - 1}{r}(r\sigma_{B2}^2 
+ (2r - 1)\sigma_{B1}^2 + (1 - r)\sigma_{J}^2 - 2(2r - 1)q\sigma_{B1}\sigma_{B2}).
\end{align}
\textcolor{black}{
As described above, we assume that task 1 has been sufficiently trained so that $\epsilon_{g1}$ is close to ``0,'' so we consider $B_1^i = J^i$ when $1\le i \le rN$.
}
So we set the initial value of $R_{12}^1(t)$ as $R_{12}^1(0) = (2r - 1)\sigma_{B1}^2$, the initial value of $R_{12}^2(t)$ as $R_{12}^2(0) = (2r - 1)q\sigma_{B1}\sigma_{B2}$, the initial value of $Q_{12}(t)$ as $Q_{12}(0) = (2r - 1)\sigma_{B1}^2$, the initial value of $R_1^1(t)$ as $R_1^1(0) = r\sigma_{B1}^2$, and the initial value of $Q_1(t)$ as $Q_1(0) = r\sigma_{B1}^2$.
Substituting \eqref{eq:r1^1_task2} and \eqref{eq:q1_task2} into \eqref{eq:eg1_task2}, the generalization error of task 1 while learning task 2 can be found analytically as follows:
\begin{align}
\epsilon_{g1} =\frac{\sigma_1^2}{2}\Bigl\{C_1 + (C_1 - C_2)\exp(-2t\eta r\sigma_2^2)
+ C_2\exp(-t\eta r\sigma_2^2(2 - \eta r\sigma_2^2))- 2 C_1\exp(-t\eta r\sigma_2^2)\Bigr\}.\label{eq:eg1_task2}
\end{align}

\section{Results}
\begin{figure}
\centering
    \subfigure[The learning curve of $N= 3000, r = 0.8, q = 0.3, \sigma_{B1} = \sigma_{B1} = \sigma_{J} = 1, \sigma_1^2 = \sigma_2^2 = 0.8$.]{\includegraphics[width = 13cm]{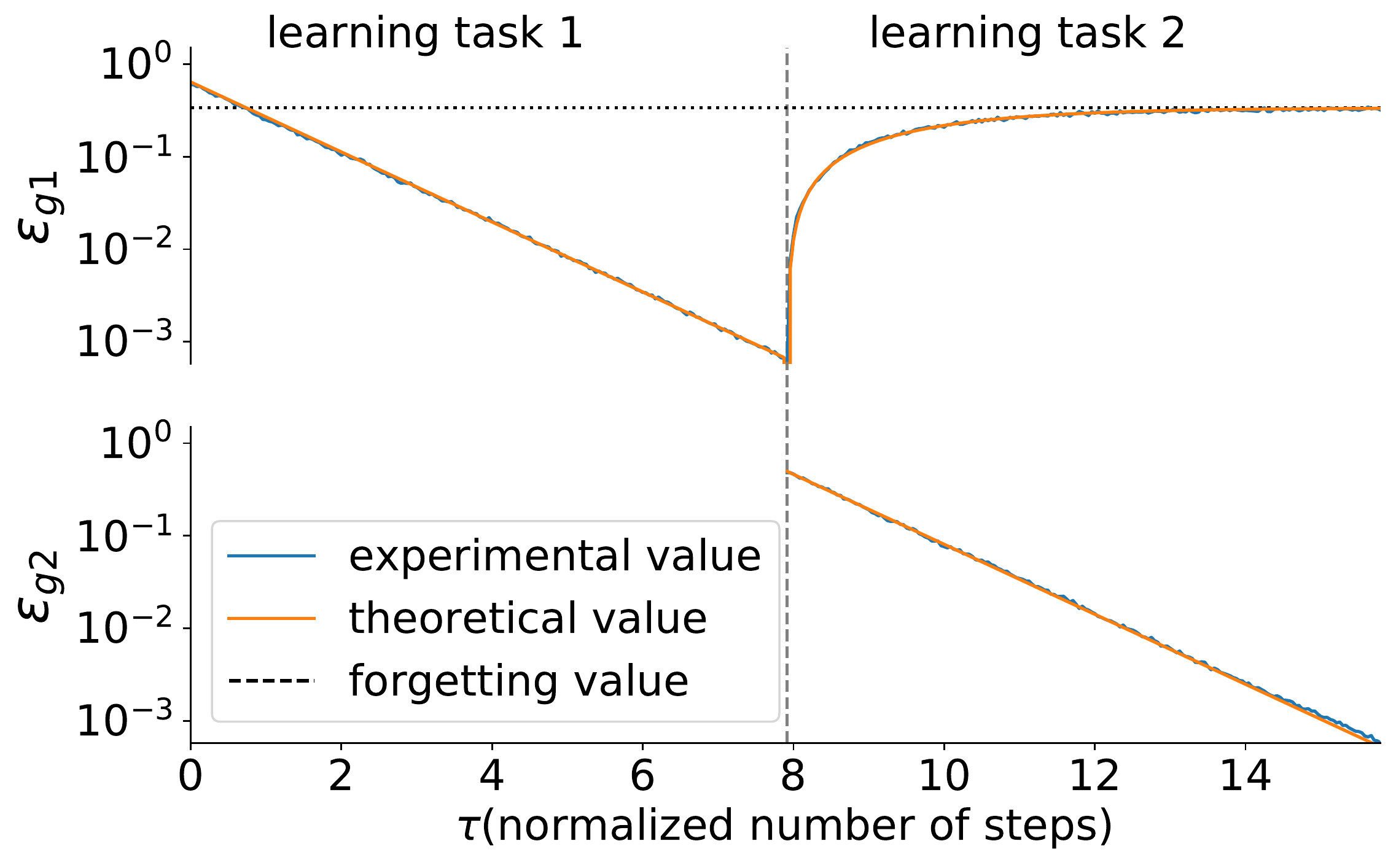}\label{fig:result_1}}\\
    \subfigure[The learning curve of $N= 3000, r = 0.8, q = 0.9, \sigma_{B1} = \sigma_{B1} = 1, \sigma_{J} = 11, \sigma_1^2 = \sigma_2^2 = 1.7$.]{\includegraphics[width = 13cm]{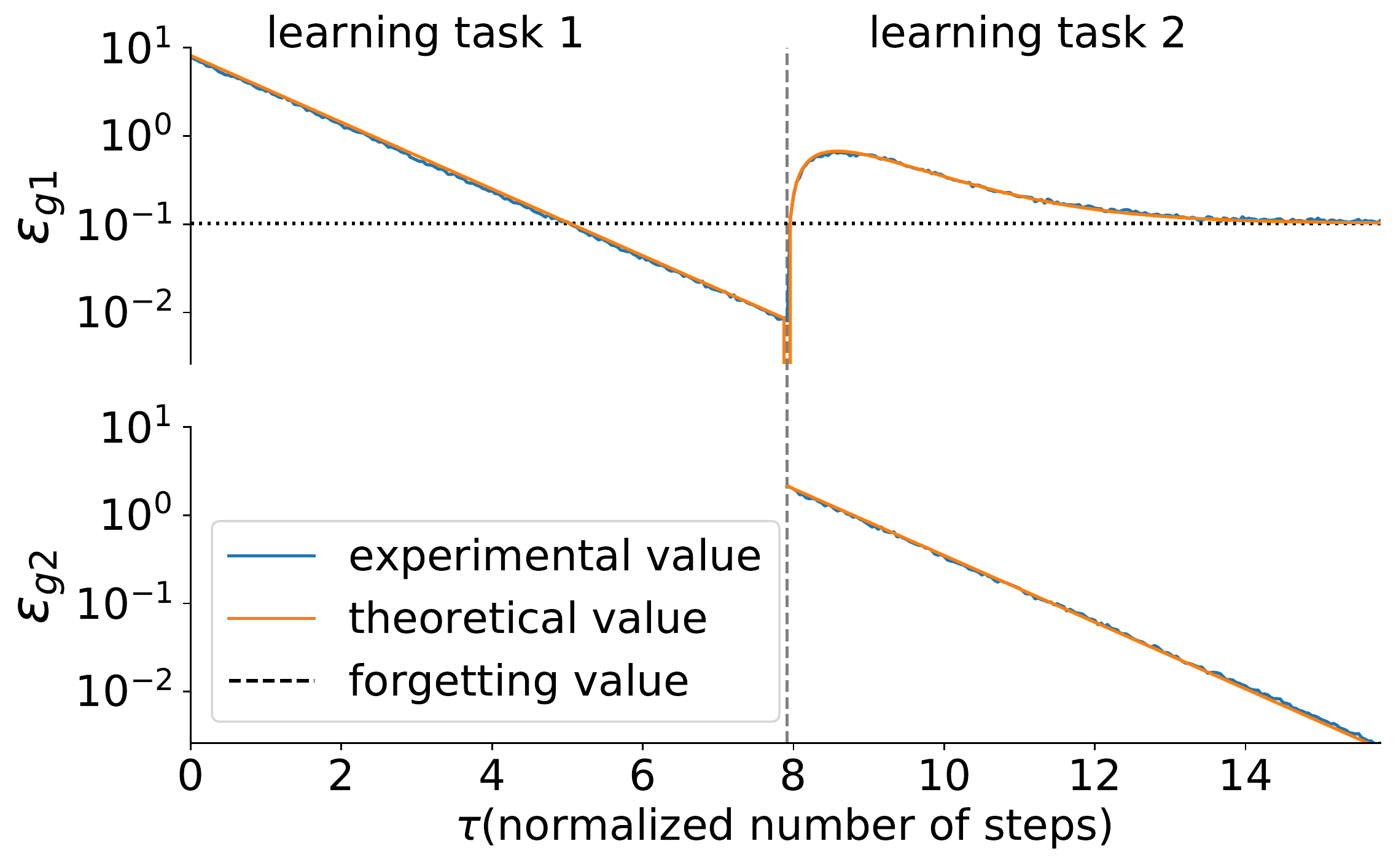}\label{fig:result_2}}
\caption{The learning curve. Blue lines indicate experimental values, and orange lines indicate theoretical values. In both figures, the experimental and theoretical values overlap very well. Two different behaviors can be observed depending on the parameter settings. We will discuss this phenomenon in Section \ref{subsec:overshoot}.
}
\end{figure}

In this section, we verify that the theoretical values of the generalization errors \eqref{eq:eg1_task1}, \eqref{eq:eg2_task2} and \eqref{eq:eg1_task2}, as well as the experimental values of the generalization errors, are consistent. We discuss the behavior of generalization errors.
Since the generalization error of task 1 during task 2 learning showed two typical behaviors, which is discussed in Section \ref{subsec:overshoot}, we present two experimental results in Figs. \ref{fig:result_1} and \ref{fig:result_2}.
Fig. \ref{fig:result_1} shows the learning curve of $N= 3000, r = 0.8, q = 0.3, \sigma_{B1} = \sigma_{B1} = \sigma_{J} = 1, \sigma_1^2 = \sigma_2^2 = 0.8$, and Fig. \ref{fig:result_2} shows the learning curve of $N= 3000, r = 0.8, q = 0.7, \sigma_{B1} = \sigma_{B1} = 1, \sigma_{J} = 2, \sigma_1^2 = \sigma_2^2 = 1.7$.
Blue lines indicate experimental values, and orange lines indicate theoretical values.
In both Figs. \ref{fig:result_1} and \ref{fig:result_2}, the experimental and theoretical values overlap very well, and we can see that the theory we have derived in \eqref{eq:eg1_task1}, \eqref{eq:eg2_task2} and \eqref{eq:eg1_task2} is valid.

As can be determined from Figs. \ref{fig:result_1} and \ref{fig:result_2}, two different behaviors can be observed depending on the parameter settings. 
In Fig. \ref{fig:result_1}, the generalization error approaches the limit of $\epsilon_{g1}$ from below.
In Fig. \ref{fig:result_2}, the generalization error greatly exceeds the limit of $\epsilon_{g1}$ once and then converges.
We refer to the phenomenon in Fig. \ref{fig:result_2} as \textit{overshoot}. In Section \ref{subsec:overshoot}, we will discuss overshoot in detail.
We also refer to the limit of the generalization error of task 1 after learning task 2 as forgetting value $\epsilon_{g1}^\prime$.
This is because $\epsilon_{g1}$ approached $0$ immediately after the completion of task 1 learning but rose to $\epsilon_{g1}^\prime$ after the completion of task 2 learning, indicating that task 1 was forgotten by learning task 2. We will discuss the forgetting value in Section \ref{subsec:forgetting_value}.

\subsection{Forgetting value}\label{subsec:forgetting_value}
In this section, we discuss the forgetting value.
The forgetting value is the limit of the generalization error of task 1 while learning task 2, so we can obtain forgetting value $\epsilon_{g1}^\prime$ in \eqref{eq:eg1_task2}:
\begin{align}
\epsilon_{g1}^\prime = \frac{\sigma_1^2}{2}(2r - 1)(\sigma_{B1}^2 + \sigma_{B2}^2 - 2q\sigma_{B1}\sigma_{B2}).\label{eq:forgetting_value}
\end{align}

\begin{figure}[tb]
\centering
\includegraphics[width = 10.0cm]{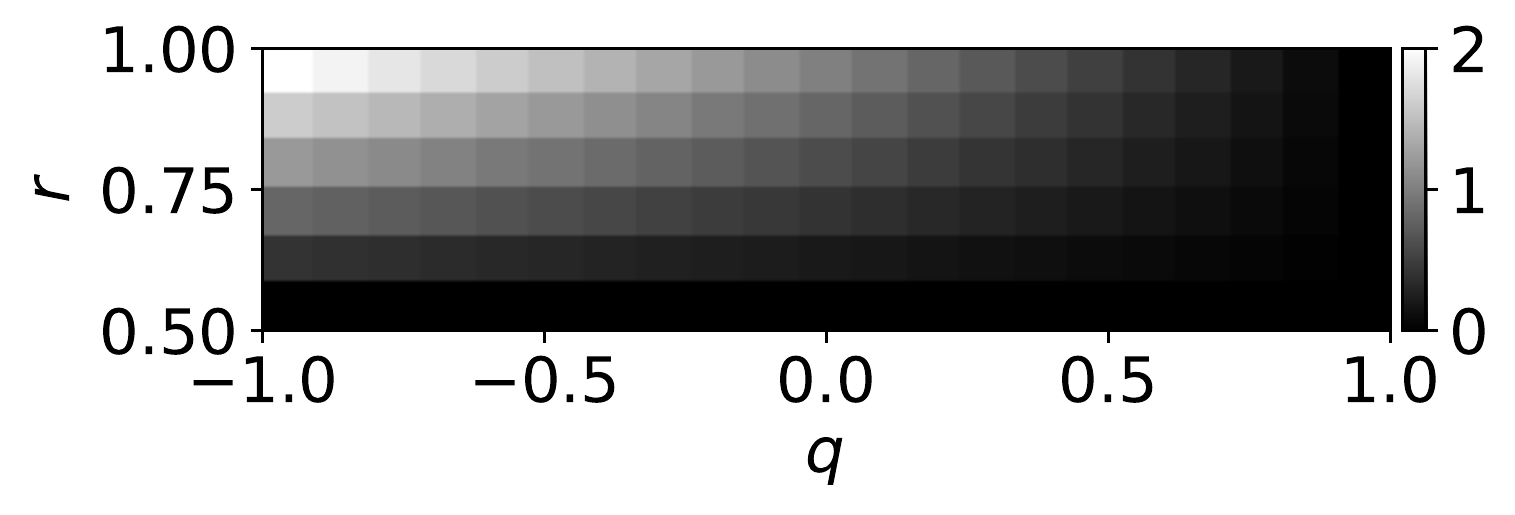}
\caption{Heat map showing the limit of the generalization error of task 1 while learning task 2. Let $r, q$ be variables and $N = 1500,\sigma_{B1}^2 = \sigma_{B2}^2 =  \sigma_J^2 = 1 $.}
\label{fig:heatmap}
\end{figure}

To determine the contributions of input space similarity $r$ and weight space similarity $q$ to $\epsilon_{g1}^\prime$, we generate a heat map of forgetting values in Fig. \ref{fig:heatmap}. 
We numerically calculated the forgetting value according to various $r$ and $q$. 
Fig. \ref{fig:heatmap} visualizes these results with a heat map for $N = 1500,\sigma_{B1}^2 = \sigma_{B2}^2 = \sigma_J^2 = 1$.
The darker color represents smaller forgetting values $\epsilon_{g1}^\prime$.

This heat map shows that when the input space similarity $r$ is small and the weight space similarity $q$ is large, the forgetting value is small. This means that the student network remembers task 1 better if $q$ is large and $r$ is small.
In other words, the student network can remember task 1 relatively well when two teacher networks are similar and two inputs are well separated.
We also found that input space similarity $r$ and weight space similarity $q$  affect the generalization error through multiplication and have an inverse effect on the forgetting value.

From \eqref{eq:eg1_task2}, we also know that the student network remembers task 1 well if $\sigma_1^2$ is small, that is, when similar inputs for task 1 are given.
Moreover, the input for task 2 is not related to the forgetting value and contributes significantly to the behavior of catastrophic forgetting.
This means that the forgetting value is dependent on the characteristics of the input for task 1, while the input for task 2 is one of the factors governing the behavior of catastrophic forgetting.

\subsection{Overshoot}\label{subsec:overshoot}
As mentioned in Section \ref{sec:introduction}, it is important to understand the behavior of catastrophic forgetting. In this section, we discuss overshoot as a characteristic behavior of catastrophic forgetting.
Overshoot is a phenomenon in which the generalization error of task 1 while learning task 2 greatly exceeds the forgetting value once and then converges.
Therefore, overshoot can be interpreted as a phenomenon in which the model forgets task 1 by learning task 2, while the model learns task 1 again through the learning of task 2. 

Here, we discuss the condition in which overshoot occurs qualitatively. \textcolor{black}{
First, we consider obtaining the condition for the overshoot analytically.
The first step is to find the time at which the time derivative of the generalization error $\epsilon_{g1}$ shown in \eqref{eq:eg1_task2} becomes zero.
The time derivative of the generalization error $\epsilon_{g1}$ can be written as follows:
\begin{align}
\dv[]{\epsilon_{g1}}{t} &= \frac{\sigma_1^2\eta r\sigma_2^2}{2}\Bigl\{-2(C_1 - C_2)\exp(-2t\eta r\sigma_2^2)\nonumber\\
&- (2 - \eta r\sigma_2^2)C_2\exp(-t\eta r\sigma_2^2(2 - \eta r\sigma_2^2))+ 2 C_1\exp(-t\eta r\sigma_2^2)\Bigr\}. \label{eq:eg1_task2_prime_t}
\end{align}
However, as can be seen from the equation, the multiplication and summation of $\exp(t(r\eta\sigma_2^2)^2)$, $\exp(-2(r\eta\sigma_2^2)t)$, and $\exp(-(r\eta\sigma_2^2)t)$ are mixed up, making it difficult to find an analytical solution. Therefore, %the following method is used to derive the conditions under which the overshoot occurs.
we derived the conditions under which overshoot occurs by comparing the coefficients of each term.}

The generalization error of task 1 while learning task 2 shown in \eqref{eq:eg1_task2} consists of a sum of terms in $\sum_i C_i\exp(-\alpha_i t)$. The smaller the $\alpha_i$ is, the slower the decay and the longer the contribution to the generalization error remains. Therefore, considering the signs of the coefficients of the terms with the smallest $\alpha_i$, we can determine whether overshoot occurs.
Overshoot occurs when the coefficient of the term with the smallest $\alpha_i$ is positive. This is because $\epsilon_{g1}$ converges to the forgetting value from a value greater than $\epsilon_{g1}^\prime$.
Conversely, overshoot may not occur when the coefficient of the term with the smallest $\alpha_i$ is negative. This is because $\epsilon_{g1}$ converges to a destination from a value smaller than $\epsilon_{g1}^\prime$.
The smallest $\alpha_i$ is determined from the setting of $\eta$, $r$, and $\sigma_2^2$, and the specific conditions for these parameters regarding the occurrence of overshoot are as follows:
\begin{align}
    \text{an overshoot } \begin{cases}
        \text{ may not occur} & (0 < \eta r \sigma_2^2 < 1) \\
        \text{ does not occur} & (\eta r \sigma_2^2 = 1 \text{ and } C_2 - 2C_1 < 0) \\
        \text{ occurs} & (\eta r \sigma_2^2 = 1 \text{ and } C_2 - 2C_1 > 0) \\
        \text{ occurs} & (1 < \eta r \sigma_2^2 < 2)
    \end{cases}.
\end{align}
If $\eta r \sigma_2^2$ exceeds two, then training diverges.
Note that when $\eta r \sigma_2^2 = 1$, the dominating coefficient $C_i$ corresponds to $C_2 - 2C_1$, and its sign depends on the other hyperparameters.
Therefore, under condition $\eta r \sigma_2^2 = 1$, overshoot occurs only when $C_2 - 2 C_1 > 0$ and does not occur when $C_2 - 2 C_1 < 0$.
Additionally, note that when $0 < \eta r \sigma_2^2 < 1$, overshoot ``may not'' occur. This is because even if the coefficient $C_i$ on the smallest term of $\alpha_i$ is negative, there can be two extrema in the time development of the generalization error for task 1, and thus, overshoot can occur.

\section{Discussion}
\subsection{Effects of hyperparameters on catastrophic forgetting}
For practical applications, we consider whether it is possible to modify the update rule of the Stochastic Gradient Descent (SGD) to avoid catastrophic forgetting.
In the setting of our study, we know that learning converges in the range $0 < r\eta\sigma_2^2 < 2$. In this section, we will limit our discussion within this range.
In the case of SGD, the only parameter that can be adjusted is $\eta$, and adjusting $\eta$ does not change the amount of the forgetting value showed in \eqref{eq:forgetting_value}.
However, the speed of forgetting shown in \eqref{eq:eg1_task2_prime_t} depends linearly on $\eta r\sigma_2^2$. This means that the speed of forgetting can be reduced by making $\eta$ smaller.
On the other hand, there are proposals to improve the learning algorithms themselves, such as Elastic Weight Consolidation (EWC) proposed by Kirkpatrick et al.\cite{Kirkpatrick3521}, so there is a possibility that the framework proposed in this paper can be used to analyze these algorithms in the future.

\subsection{Overparametrization}\label{subsec:overparametrization}
In cutting-edge theoretical research, there is interest in situations where the number of student's parameters is greater than the number of teacher's parameters. We call this situation ``overparametrization.'' In this section, we discuss such a setting.
In this paper, both the student and the teacher networks are single-layer linear neural networks, so considering the extension to overparametrization, the natural situation is that the student has more input elements $N$ than the teacher. However, since teacher-student learning is not possible in such a situation, overparametrization in this setting is non-trivial.
Therefore, we will discuss the possible effects on applying the case of overparametrization using online learning and statistical mechanical analysis to continual learning.

The overparametrization in on-line learning and statistical mechanical analysis has been discussed by Goldt et al. \cite{Goldt2020HMM} and Goldt et al. \cite{Goldt19nips}.
The former case assumes a more realistic situation of a so called hidden manifold model for inputs.
The inputs from real world are considered to be concentrated on a low dimensional manifold.
Therefore, in the hidden manifold model, the inputs to the student are generated using variables sampled on a low-dimensional manifold. In this case the difference in the number of parameters does not affect the generalization error. 
We suppose that this is because the dimension of the manifold characterising the input data is smaller than the number of parameters of the student.
In the latter case, they consider a situation where the inputs are assumed to be sampled i.i.d. from a Gaussian distribution and there is noise in the output of the teacher network. In this case, they report that the convergence of the generalization error depends linearly on the difference between the number of student and teacher parameters. 
This may be because the weights of the students contribute equally to the learning due to the high symmetry of the input, so that the weights of the students that are not needed to learn the teacher learn the noise of the teacher's output.

Assuming that these results are qualitatively the same in a continual learning framework, we can expect the following to happen.
In the former case, since the difference in the number of parameters does not affect the generalization error, we expect the results on catastrophic forgetting to be qualitatively unchanged.
On the other hand, in the latter case, the generalization error increases as the number of student parameters increases. 
Since forgetting means that the generalization error, once reduced, increases, we expect that the apparent forgetting will be small if the error has not been reduced originally.
Therefore, in this case, overparametrization may produce qualitatively different results.

\subsection{In the case of input sampled from a non-Gaussian distribution}
In the real world, the inputs are generally sampled from a non-Gaussian distribution.
In this section, we discuss how the implication changes when the input distribution is sampled from a non-Gaussian distribution.

Goldt et al.\cite{Goldt2020HMM} described in Section \ref{subsec:overparametrization} discusses the situation of online learning and statistical mechanics analysis when the inputs are not sampled from a Gaussian distribution but a general one.
In the hidden manifold model, the input is assumed to be sampled from a low-dimensional manifold rather than a Gaussian distribution. The dimension of the manifold is called the intrinsic dimension.
As explained above, in the hidden manifold model, the generalization error of the convergence destination no longer depends on the learning rate or the difference between student and teacher parameters. Instead, the ratio of the input dimension to the intrinsic dimension influences the generalization error.
We will discuss possible influences of this consequence on our two findings.
One finding is that the forgetting value is characterized by the weight space similarity and the input space similarity. The other is the phenomenon called overshoot.
The following is a discussion of how these findings may change when the inputs are extended to non-Gaussian in a setting such as the hidden manifold model described above.

First, let us consider the possible impact on the first finding.
Input space similarity quantifies the degree to which the input data overlap in the input dimension. As mentioned above, in a hidden manifold model, the intrinsic dimension of the manifold that generates the data is more important than the apparent input dimension.
Therefore, in addition to the input space similarity, the similarity of the intrinsic dimensions of the manifold may be important.
%if the input space similarity is high, but the similarity on the manifold is low, the similarity of the input will not affect forgetting, and the similarity on the manifold will be more important.

Next, we consider the effect on the second finding, that of overshoot.
Goldt et al.\cite{Goldt2020HMM} reported that learning proceeds more quickly without being trapped by poor local solutions under the assumption of a hidden manifold model.
An overshoot is a phenomenon where the learning rate is too large and learning proceeds too fast, temporarily increasing the generalization error of old tasks than that at convergence destination.
If we assume that learning is also accelerated in continual learning by making the input distribution non-Gaussian, this is equivalent to a larger learning rate in the execution, and overshoot may become more pronounced. In this case, the conditions under which overshoot occurs may be modified.

\section{Conclusions}
In this study, we provided a theoretical framework for analyzing catastrophic forgetting by using teacher-student learning.
We theoretically analyzed the generalization error of the student network for continual learning of two tasks.
We found that input space similarity and weight space similarity exerted significant effects on catastrophic forgetting.
The student network can remember the first task relatively well when the similarity of the input distribution is small and when the similarity of the input-output relationship is large.
We also find that a characteristic phenomenon, which we refer to as \textit{overshoot}, is observed under the specific hyperparameter conditions.
This phenomenon suggests that even if a system undergoes catastrophic forgetting, it may still exhibit better performance if it learns the current task for a longer period of time.

\begin{acknowledgment}
%\acknowledgment
% For environments for acknowledgement(s) are available: \verb|acknowledgment|, \verb|acknowledgments|, \verb|acknowledgement|, and \verb|acknowledgements|.
This work was supported by JSPS KAKENHI (Scientific Research (A)18H04106) and by JST PRESTO (JPMJPR17N2).
\end{acknowledgment}

\bibliography{cite}
\bibliographystyle{jpsj}

\end{document}